# Ontology-Enhanced Educational Annotation Activities

**Joaquín Gayoso-Cabada[1], María Goicoechea-de-Jorge[2], Mercedes Gómez-Albarrán[3], Amelia Sanz-Cabrerizo[4], Antonio Sarasa Cabezuelo[5], José-Luis Sierra-Rodríguez[6,*]**

[1] Fac. Informática, Universidad Complutense de Madrid; jgayoso@ucm.es
[2] Fac. Filología, Universidad Complutense de Madrid; mgoicoe@ucm.es
[3] Fac. Informática, Universidad Complutense de Madrid; mgomeza @ucm.es
[4] Fac. Filología, Universidad Complutense de Madrid; amsanz@ucm.es
[5] Fac. Informática, Universidad Complutense de Madrid; asarasa @ucm.es
[6] Fac. Informática, Universidad Complutense de Madrid; jlsierra@ucm.es
**\*** Correspondence: jlsierra@ucm.es; Tel.: +34-91-3947548

**Abstract:** The educational potential of digital document annotation tools is widely recognized. Using these tools, students can enrich the documents with annotations that highlight the most relevant aspects of these documents. As the conceptual complexity of the learning domain increases, the annotation of the documents may require comprehensive domain knowledge and an expert analysis capability that students usually lack. Consequently, a proliferation of irrelevant, incorrect and/or poorly decontextualized annotations may appear, while other relevant aspects are completely ignored by the students. The main hypothesis proposed by this paper is that the use of a guiding annotation ontology in the annotation activities is a keystone aspect to alleviate these shortcomings. Consequently, comprehension is improved, exhaustive content analysis is promoted, and meta-reflective thinking is developed. To test this hypothesis, we describe our own annotation tool, *@note*, which fully implements this ontology-enhanced annotation paradigm, and we provide experimental evidence about how *@note* can improve academic performance via a pilot study concerning critical literary annotation.

## 1. Introduction

Annotation activities have high value in the learning process; in recent years, they have taken advantage of digital document annotation tools. These tools enable learning activities based on the addition of marks and comments in the contents of the documents, which enable highlighting, complementing and enriching some of their aspects. However, as the conceptual complexity of the learning domains increases, annotation tasks require a comprehensive knowledge of the domain [1]. Consequently, annotation activities are difficult for students who frequently highlight irrelevant aspects of the content in some of their annotations and bypass other aspects of utmost importance to satisfy a certain annotation objective. As a result, learning outcomes are suboptimal. Thus, guidance is required to provide a better learning experience. Unfortunately, most of the available annotation tools focus on providing appealing user experiences (e.g., enrichment by multimedia contents) or supporting collaborative annotation by groups of students more than providing adequate guidance for achieving successful educational experiences [2].

In this paper, our main research hypothesis is that the use of suitable *annotation ontologies* can provide students with the necessary guidance during the annotation of documents in complex learning domains. Instructors provide annotation ontologies that capture their particular conceptual frames and annotation goals. Students benefit from a document annotation paradigm that properly combines semantic and free-text annotation approaches. Thus, students can express the various aspects of the analyzed documents in natural language while using a guiding ontology to properly classify their annotations. This explicit classification of annotations promotes meta-reflective thinking since students are forced to reflect on the purpose of every annotation in the context of the activity. Our annotation paradigm has been put into practice in the *@note* annotation tool, whose initial conception and some subsequent evolutions are described in [3–5].

The content of this paper is organized as follows. Section 2 discusses related work. Section 3 briefly describes the methodology used to study the effect of annotation ontologies in the annotation process, introducing @note and describing a pilot study concerning *critical literary annotation*. Section 4 describes the results obtained in the pilot. Section 5 provides a discussion of these results. Section 6 concludes the paper and presents future work.

## 2. Related work

Unlike other works related to annotation tools, which emphasize aspects such as comparing digital versus handwritten annotations [6], analyzing how digital annotations impact the reading process [7,8] or how readers can become actively involved in the text annotation process, paying special attention to collaborative annotation of texts by groups of annotators [9–14], in this paper we mainly focus on the mechanisms offered by the annotation tools for classifying annotations. For this purpose, we identified five main approaches concerning the classification of annotations (see [15] for a more detailed account):

- *Lack of mechanisms for classifying annotations*. This approach groups tools that prioritize other aspects instead of organizing annotations, such as sophisticated ways of interacting with the documents.
- *Predefined annotation modes*. Tools that follow this approach provide different ways of annotating a document (underlining or highlighting fragments and adding comments), which induce an implicit categorization of the annotations.
- *Pre-established semantic categories*. This approach is adopted by tools that introduce a predefined set of semantic tags to classify annotations.
- *Folksonomies*. This approach is based on the classification of annotations by mean of tags that are created by the users to conform a folksonomy. Previously created tags (by the user or other participants in the annotation activity) can be employed or new tags that are better suited to particular classification needs can be created.
- *Ontologies*. Tools that adhere to this approach enable loading specific ontologies for each annotation activity. Students can use these ontologies to make the semantics of annotations explicit (e.g., by associating one or more concepts in the ontologies to the annotations).

Table 1 provides representative examples of tools in each of the categories. Concerning the guidance of students during the annotation process in complex domains:

- Tools that lack mechanisms for classifying annotations and tools based on predefined annotation modes imitate conventional paper-and-pencil-based annotation mechanisms, perhaps modulated with a greater repertoire of presentation styles. Therefore, they do not provide any mechanism to support students' guidance.
- Concerning tools that are based on predefined repertories of semantic categories, although the lists of tags provided by these repertories allow annotations to be classified according to certain semantic criteria, their pre-established nature produces generic classification systems, which typically consist of general purpose and reduced sets of universal categories (4 in *PAMS 2.0* or in *MyNote*, 7 in *CRAS-RAID*, 9 in *Tafannote* or in *MADCOW*, etc.), which may not fit the specific characteristics of every annotation activity.
- Folksonomy-based tools, enable lists of semantic tags that are specifically adapted to each annotation activity. However, the majority of these tools delegate the collaborative design of these vocabularies of tags to the students. For instructors, this practice does not guarantee that the resulting folksonomies adequately capture the objectives of the annotation activity, because it requires a considerable amount of expert knowledge that students lack. Although some of the folksonomy-based tools (e.g., *annotation studio*) also provide support for tag repertories provided by instructors, folksonomies lack structure beyond the provided by simple tag lists, which can be inconvenient for in-depth annotation.

Table 1. Examples of tools categorized according to their annotation classification approaches.

| **Annotation classification approach** | **Examples of Tools** |
|---|---|
| Absence of classification mechanisms | *Digital Reading Desk* [16] |
| | *Livenotes* [17] |
| | *WriteOn* [18] |
| | *PaperCP* [19] |
| | *u-Annotate* [20] |
| Predefined annotation modes | *Adobe Reader*[1] |
| | *PDF Annotator*[2] |
| | *Diigo*'s[3] annotation tool [21] |
| | *Amaya*'s annotation tool [22], *Anozilla*[4] |
| | *CASE* [23] |
| | *CON2ANNO* [24] |
| | *Online annotation system* [25] |
| | *VPen* [10] |
| | *IIAF* [26]. |
| Pre-established semantic categories | *eLAWS* and *Annoty* [27] |
| | *Highlight* [28] |
| | *PAMS* 2.0 [14] |
| | *MyNote* [29] |
| | *Tafannote* [30] |
| | *WCRAS-TQAFM* [8] |
| | *CRAS-RAID* [13] |
| | *UCAT* [31] |
| | *MADCOW* [32] |
| Folksonomies | *HyLighter*[5] [33] |
| | *Hypothe.sis*[6] [34,35] |
| | *annotation studio* [36] |
| | *A.nnotate* [37,38] |
| | *Note-taking* [39] |
| | *OATS* [40,41] |
| | *SpreadCrumbs* [6,42] |
| | *Tsaap-Notes* [43] |
| Ontologies | *Loomp* [44,45] |
| | *DLNotes* [46] |
| | *MemoNote* [47,48] |
| | *WebAnnot* [49] |
| | *New-WebAnnnot* [50] |

- Ontology-based tools provide appropriate vehicles (ontologies) for capturing specific knowledge about annotation activities, which enables the adaptation of the tools to the semantic particularities of each activity and provide a high degree of contextualization in this activity. In addition, the structural richness of ontologies solves the problems of lack of structure of plain lists of semantic tags.

---

[1] acrobat.adobe.com

[2] www.pdfannotator.com

[3] www.diigo.com

[4] annozilla.mozdev.org

[5] www.hylighter.com

[6] web.hypothes.is

From these annotation classification approaches, the approaches that are based on explicit ontologies appeared as the most appropriate ones to guide students. However, the tools in this category have some shortcomings, which hinder their applicability:

- The complexity of the ontology definition by instructors must be carefully considered. Of the tools that were analyzed, only *Loomp* addresses this aspect; it proposes a two-level organization scheme that is based on *vocabularies* that cluster atomic concepts. This approach is too simple for conceptual organization purposes. The other tools adopt standard semantic web technologies (like RDFS or OWL) and do not introduce mechanisms to help instructors provide the ontologies.
- All ontology-based tools that were analyzed differentiate between semantic annotations and other types of annotations. This fact is evident, for example, in *DLNotes*, which explicitly distinguishes between semantic annotations and free-text annotations. The other tools focus on the process of semantic annotation, which is understood as semantic tagging of document fragments. From a detailed annotation perspective, providing textual content to annotations in free-text format is essential to reflect the particular and subjective reading of the content by the student.

## 3. Materials and Methods

To study how the emphasis on annotation ontologies can enhance educational annotation activities in complex domains, we adopted a methodology that is grounded on *design-based research methods* [51]:

- Following the guidelines of design-based research methods, to address the aforementioned shortcomings of ontology-based annotation tools we designed and developed our own annotation tool, *@note*, which fully implement our annotation paradigm. This tool is detailed in subsection 3.1.
- To assess the educational utility of *@note*, we undertook a pilot experiment concerning a learning domain that requires a large amount of domain knowledge and skilled annotation capabilities: *critical literary annotation*. Concerning the quantitative analysis method in this experiment, we opted for a *within-subject design* approach [52] since it fits reduced groups of students, which typically arise in advanced university-level literature courses. The pilot design is detailed in subsection 3.2.

*3.1. @note annotation tool*

Being aware of the shortcomings for the educational use of ontology-based annotation tools, with *@note* we made a strong emphasis on integrating free-text and semantic approaches in a single annotation paradigm. Additionally, we also doted *@note* with user-friendly ontology-edition mechanisms suitable for end users with no specific background on computer science or knowledge engineering. The central construct of *@note* is that of *annotation activity*. These activities are defined by instructors and they can be carried out by students. Thus, the tool offers instructors a complete set of features to design annotation activities (paragraph 3.1.1). Amongst these features, the most relevant ones are those oriented to the provision of annotation ontologies (paragraph 3.1.2). Once the annotation activity has been properly customized, *@note* provides students with an interactive environment in which to make annotations (paragraph 3.1.3) and instructors with several features for assessing the annotation activities (paragraph 3.1.4).

*3.1.1. Annotation activities*

*Annotation activities* in *@note* are oriented to accomplish specific annotation objectives (what aspects should be annotated, and for what purpose). *@note* facilitates instructors in conceiving and refining these activities by providing them with user-friendly editing tools for configuring each of these aspects. Each annotation activity comprises the following aspects:

- The *document to annotate*. When designing an annotation activity, the instructor must choose the document to be annotated by the students. In the current version of *@note*, these documents can be obtained from the collection of Google digitized books (Google Books[7]) or by directly loading documents into the tool (Figure 1a).
- The *annotation group*. This set of students will be in charge of performing the activity (Figure 1b).
- The *annotation ontology*. This ontology will guide the students throughout the annotation process (Figure 1c).

The most critical aspect is the one that is concerned with the annotation ontologies. This critical aspect is discussed in the next paragraph.

a)

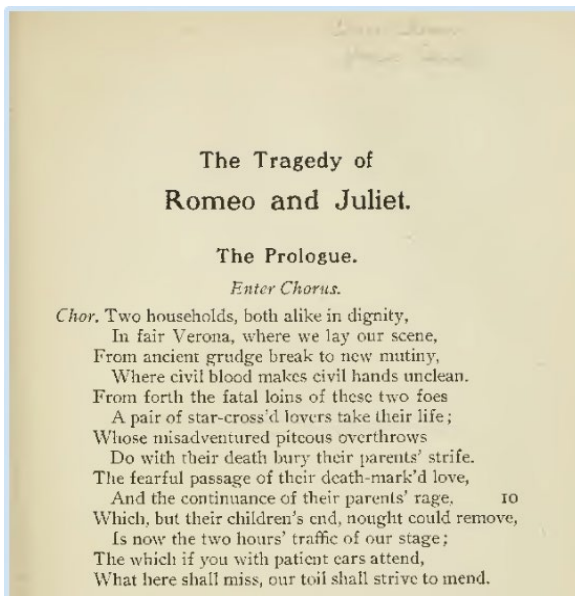

b)

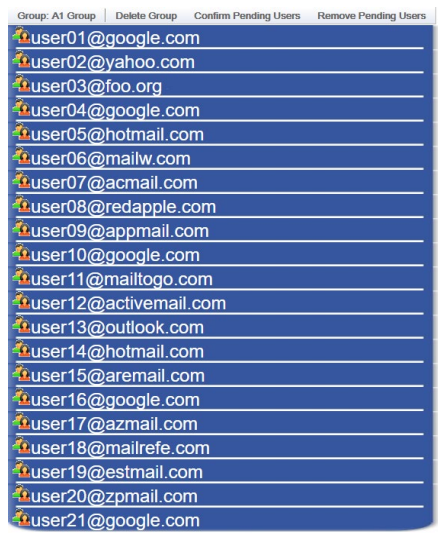

c)

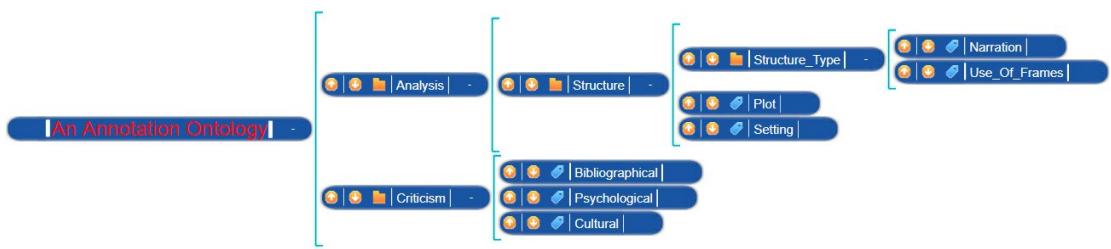

Figure 1: Snapshots concerning the creation of an annotation activity in @note for the *literary theory and text analysis* discipline: a) the document to be annotated uploaded in @note; b) the annotation group built with the group creation tool (usernames of students intentionally hidden); c) an excerpt of the annotation ontology.

*3.1.2. Annotation ontologies*

*Annotation ontologies* enable students to explicitly classify their annotations. An annotation ontology will depend on the annotation objectives of the activity. The instructor will be able to focus on different concepts based on the aspects of the content that he or she wishes to focus on during the annotation process. Depending on the nature and specificity of the activity, the instructor will provide an ontology that is oriented to the activity or he/she will profit from reusing or adapting an existing ontology. To reduce the workload of defining the annotation ontologies, *@note* enables instructors to control the visibility of their ontologies (*private* or *public*), participate in the collaborative edition of public ontologies with their colleagues, and reuse existing ontologies for new annotation activities.

---

[7] http://books.google.com/. One of the reasons to support books in Google Books is that *@note* began to be developed in the context of two grants of the Google's Digital Humanities Award Programs (editions 2010 and 2011).

*@note* enables annotation ontologies to be defined as taxonomical arrangements of concepts. Thus, *@note* adopts a compromise solution between the simplicity of the two-level organization of vocabularies in *Loomp* and the complexity of the unrestricted ontology definition formalism (as supported by the other ontology-oriented tools). The aim was to provide an approach that is sufficiently flexible for annotation and sufficiently friendly to facilitate the definition of ontologies for instructors without expertise in computer science or knowledge engineering. This solution enables instructors to define substantially complex ontologies using the user-friendly *@note* edition features. Figure 1c shows an excerpt of an ontology for a generic theory of *critical literary analysis* [53]. *@note* annotation ontologies include two kinds of concepts:

- *Intermediate* concepts can be refined in terms of other simpler subconcepts. For instance, in Figure 1c, concepts such as *Analysis*, *Structure*, *Structure_type* or *Criticism* are intermediate concepts, and *Structure_type*, *Plot* and *Setting* are direct subconcepts of *Structure*.
- *Final concepts* are concepts that the students can use to classify annotations. These concepts correspond to the taxonomy leaves. In Figure 1c, *Narration*, *Use_Of_frames*, *Plot*, *Setting*, *Bibliographical*, *Psychological* and *Cultural* are final concepts that students can use to classify their annotations.

This hierarchical organization of concepts facilitates, on the one hand, the search for the most appropriate final concepts for each annotation during the annotation of the document and, on the other hand, the filtering of the annotations during the assessment of the annotation activity, because it is possible to filter by both final concepts and intermediate concepts (in this case, the result will be all the annotations catalogued by the final concepts that descend from the chosen one).

As evidenced by Figure 1c, in *@note*, the hierarchical relationship between concepts is expressed by brackets. This representation is useful to have a complete visual snapshot of the ontology and during text annotation and activity assessment. It is also useful for editing the ontology, as the editor lets instructors rename, delete and reorganize the hierarchical arrangement of concepts, which is an essential feature to support the continuous improvement of the annotation activities.

*@note* also enables instructors to specify intermediate concepts which allows students to scope their own final concepts. This feature leaves the ontology open to the suggestions of the students. As a result, @note accommodates folksonomy-based approaches to content annotation in a controlled manner. From the instructor perspective, this feature makes it possible to diagnose possible disadvantages of the ontology and possible problems in the assimilation of the underlying theory by students.

### 3.1.3. Making annotations

Once the annotation activity has been configured by instructors, *@note* offers the students an environment in which to perform the annotation of the document. They select parts of the document and associate the content of their annotations with these fragments. In accordance with the previously mentioned considerations, they also associate one or more ontology concepts with each annotation to classify it. Consequently, an annotation in *@note* consists of (Figure 2a):

- An *anchor*. The *anchor* is an area of the digital document associated with the activity, which students can delimit with a mouse.
- A *content*. The *content* is an unconstrained description provided by the students. In *@note*, annotation contents can contain text and substantially richer multimedia content (images, video, and audio), links to other resources, and any type of HTML5-compliant information.

a)

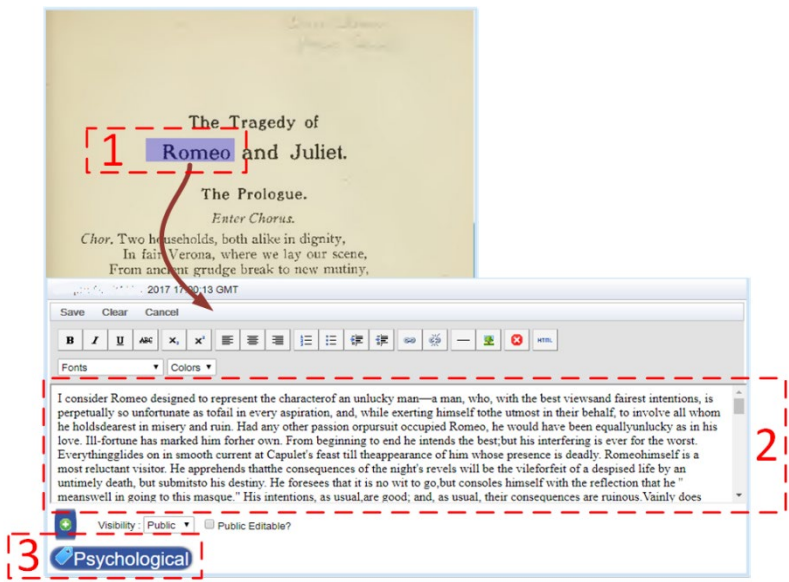

b)

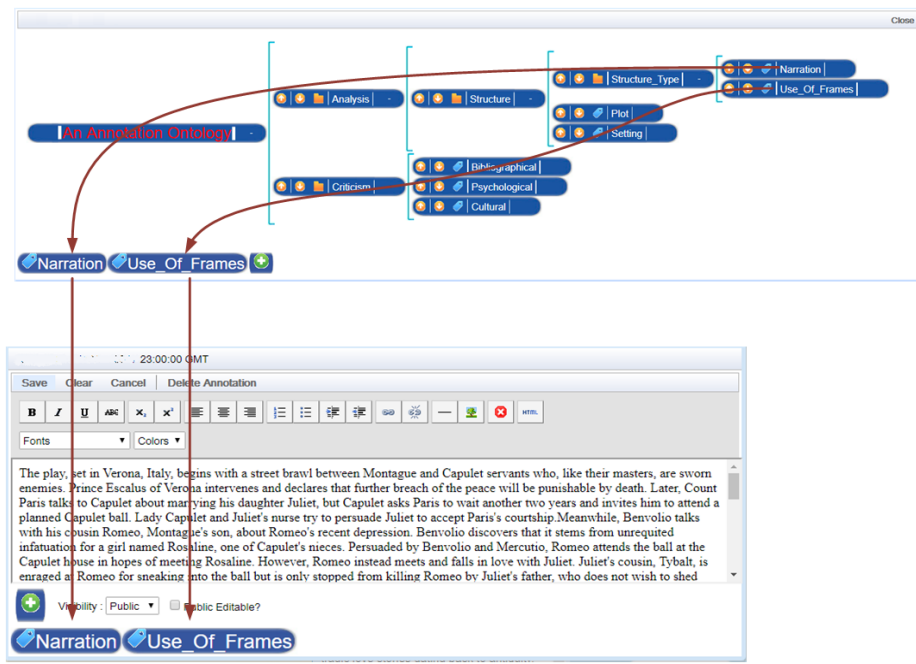

Figure 2: a) Example of a student' annotation in *@note* (1: anchor, 2: content, and 3: classification); b) semantic classification of an annotation in *@note*.

- A *classification*. The *classification* is a set of final concepts taken from the annotation ontology. Each annotation must have at least one final concept associated with it. To perform the classification of their annotations, students use the representation of the annotation ontology and select concepts that they consider to be relevant (Figure 2b). The hierarchical organization of concepts and the associated bracketed representation facilitates this task, helping students to identify the most suitable concepts and offering a complete visual snapshot of the ontology.

Thus, free-text and semantic annotation approaches are consistently combined in *@note*. Each annotation must be classified according to the annotation ontology. Likewise, transversal notes, which simultaneously address several aspects, may be semantically tagged with more than one final concept in this ontology.

*3.1.4. Assessment of annotation activities*

To support the assessment of annotation activities, *@note* offers instructors two basic mechanisms for retrieving student annotations based on their semantic classifications:

a)

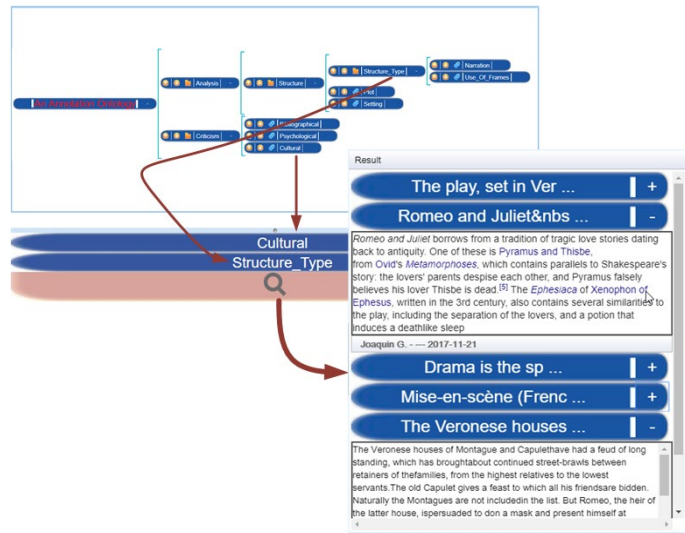

b)

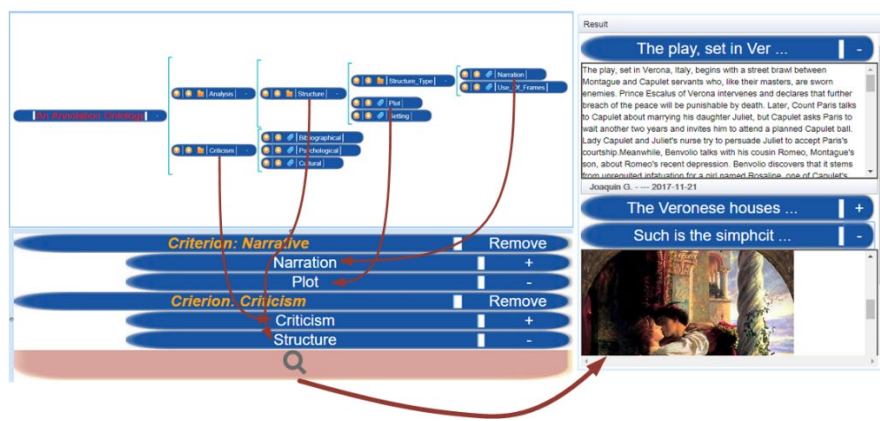

Figure 3: a) Basic filtering of annotations in @note; b) advanced filtering.

- The instructor can select all annotations that are tagged with a set of concepts, either intermediate concepts or final concepts. The annotations that are selected will be the annotations that are tagged with the final concepts chosen by the instructor and at least a final subconcept of each intermediate concept chosen by the instructor. Figure 3a illustrates this mechanism: the annotations that are selected will be the annotations that are tagged as *Cultural* and a final subconcept of *Structure_Type* (i.e., by *Narration* or *Use_Of_Frames*).
- The instructor can also use a more sophisticated search engine based on arbitrary Boolean queries in conjunctive normal form. Concepts can be asserted or denied and grouped together to express conjunctions. Each of these conjunctions is referred to as a *criterion*. Final queries are formulated by disjunctions of criteria. For example, Figure 3b illustrates the edition and application of a query with two criteria: the first criterion is named *Narrative* and enables the selection of annotations that are tagged with the *Narration* concept but are not tagged with the *Plot* concept (asserted concepts are marked with '+', while denied concepts are marked with '-');

the second criterion is named *Criticism* and selects annotations that are tagged with a subconcept of *Criticism* but are not tagged with any subconcept of *Structure*.

These functionalities are oriented to facilitate instructors in evaluating the quality of annotations. Instructors can use the annotation ontology to filter the annotations of each student, which enables the degree of compliance with the annotation objective to be assessed. Instructors can check the degree of consistency between the content of each annotation and the concepts used to classify it, which enables the extent to which the student has assimilated the conceptual structure that underlies the learning domains to be assessed. Instructors can also check if students have addressed all relevant aspects of the text or if they omitted an important aspect. This approach enables the instructors to evaluate the degree of detail and the comprehension achieved by the student during the reading of the content. These assessment facilities also promote continuous improvement of the annotation activity. Once the assessment results have been analyzed, the instructor can review his or her work and refine the critical annotation objectives and the ontology to improve student performance.

*3.2. Pilot experiment*

To assess the impact of the proposed annotation paradigm in the students' learning performance, we undertook an experiment in the domain of *critical literary annotation*. *Critical literary annotation* is one of the basic competencies acquired in any literary program at the university level. In critical literary annotation, students describe how a literary text is built and functions. Annotations are typically linked to text fragments and can exhibit various types: explanation of historical context, linguistic aspects, semantic aspects, influences, literary style, etc. [1].

We framed our experiment in a French literature course in a French studies degree program. The pilot involved 28 students who were enrolled in the course. As previously mentioned, this relatively small number of students in a typical advanced course of a university-level degree program in literature, compelled us to adopt a *within-subject* approach. Contrary to typical (*between-subject*) experimental designs that are based on a control group and experimental group, in a *within-subject* design, every participant is subjected to every treatment, which significantly contributes to the reduction of the variance and the number of required participants. Its main disadvantage is the potential presence of carryover effects (e.g., fatigue or practice), which can thread internal validity. In our case, we estimated the impact of these carryover effects in the final results to be negligible (section 5). The pilot ran as follows:

- After reviewing the key principles of structural and thematic narratology for the analysis of narrative texts, the students were instructed in the conventional practice of annotation with paper and pencil [54] to initiate a narrative-type analysis.
- Then, they were asked to annotate a first text obtained from *Les Liaisons dangereuses* by eighteenth-century French author Choderlos de Laclos [8]. This activity was employed as a baseline in the *within-subject* design. Twenty-six of the 28 students participated in this activity (two absences were recorded). Students worked during a one-hour class session and were given the option to finalize the activity during the following week.

---

[8] Choderlos de Laclos, *Les liaisons dangereuses*, Amsterdam, Durand, 1784.

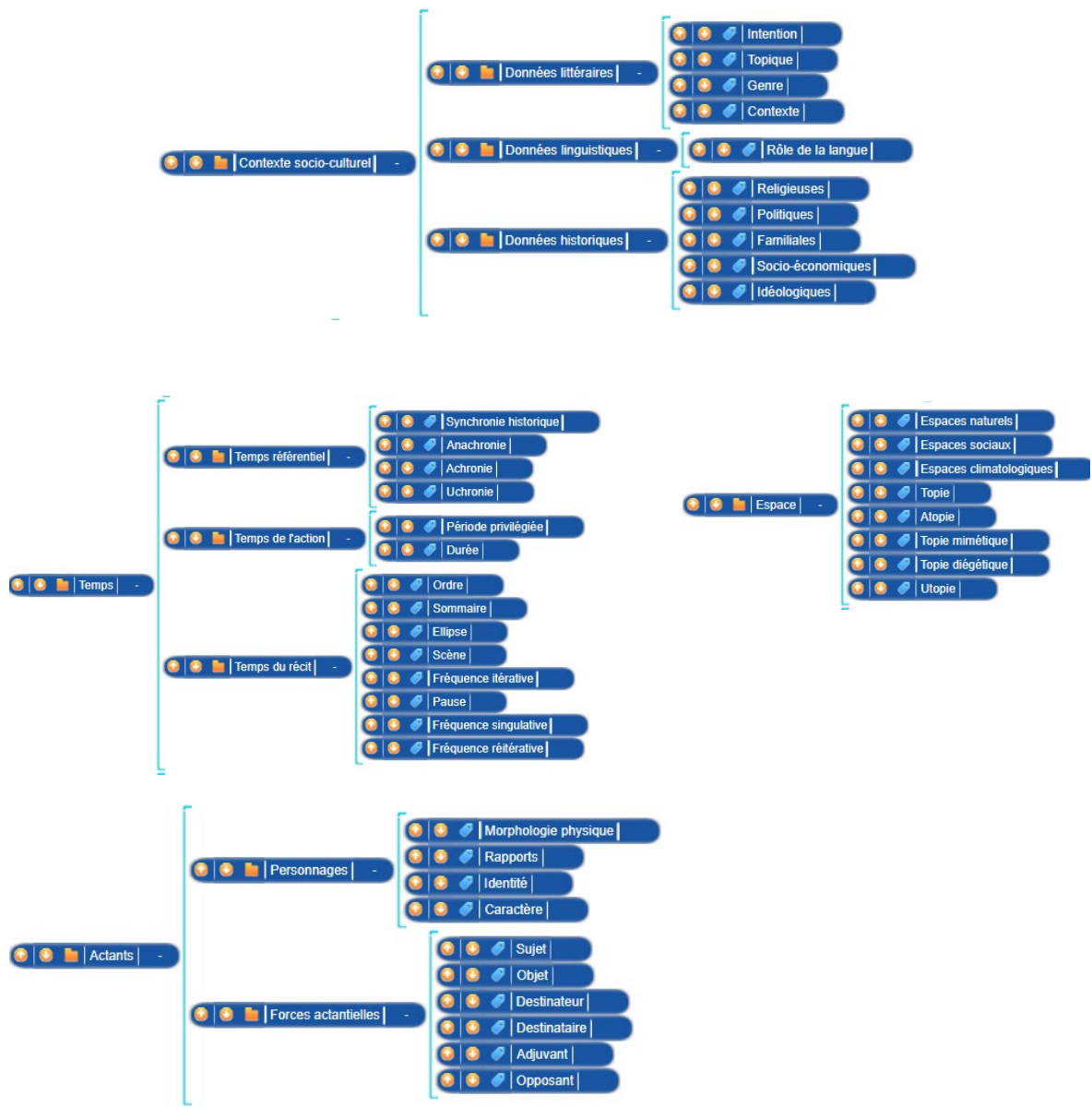

Figure 4: Annotation ontology for *Les Liaisons dangereuses*

- In parallel, an annotation activity was designed in *@note* for a second text, which was also obtained from *Les Liaisons dangereuses* with a complexity that was similar to that of the baseline activity. Figure 4 outlines the annotation ontology provided by the instructor. As the ontology is aimed at students of French Studies, the instructor used French to name the concepts. This ontology includes and structures basic concepts related to the analysis of a narrative text following narratological criteria. The ontology introduces intermediate concepts of the first level to capture the main aspects contemplated in narratology (Bal, 2017): sociocultural aspects of the text (*Contexte socio-culturel* concept), aspects related to the space (*Espace*) of the narration (i.e., to the frame or place where the events occur and the characters are placed), temporal aspects of the narrative (*Temps*), actors that lead the action (*Actants*), aspects related to the author (*Auteur*), aspects related to the narrator (*Narrateur*), and aspects related to discourse analysis (*Discours*). These aspects are refined in terms of more elementary narratological concepts, as the ontology outlined in Figure 4 indicates. In this way, this ontology guides students in the process of analyzing a narrative text following very well-defined narratological criteria. The ontology is applicable and reusable in other contexts. This fact is reflected in the average number of subconcepts for each intermediate concept (4.43), which is a relatively high value that denotes the horizontal nature of the ontology (Tartir et al., 2005).
- One week after completing the paper and pencil baseline annotation activity, a one-hour session was dedicated to instruct students in the critical annotation of texts with *@note*. In the next session, students undertook the *@note* activity and worked in class for an additional hour, in which questions about the use of the tool were answered. Similar to the paper-and-paper baseline activity, the students had a week to complete the critical annotation activity with *@note*. A total of 27 students participated in this activity: 25 of the 26 students who participated in the previous (paper and pencil based) activity and 2 other students who had not attended this previous activity.

## 4. Results

This section details the results of the pilot experiment. Subsection 4.1 summarizes the results of the activity based on paper and pencil. Subsection 4.2 describes the results of the activity based on *@note*. Subsection 4.3 provides a comparison.

### 4.1. Paper and pencil based annotation

Figure 5a summarizes the distribution of the number of annotations per student for the 26 students who participated in the paper and pencil based annotation activity (a total of 468 annotations were recorded). More than 53% of the students produced between 10 and 20 annotations. The average number of annotation was 18 (95% CI [14.04, 21.96]). Once the annotations were analyzed by the instructor, the shortcomings expected of an unguided annotation activity formulated in a complex domain such as *critical literary analysis* were detected:

(a) (b)

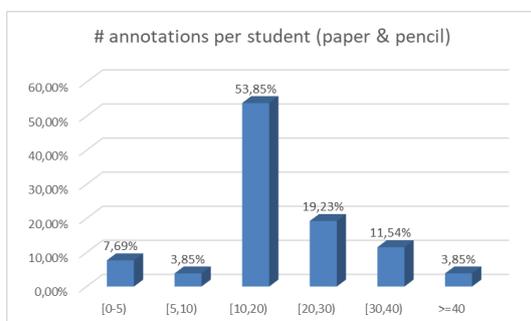 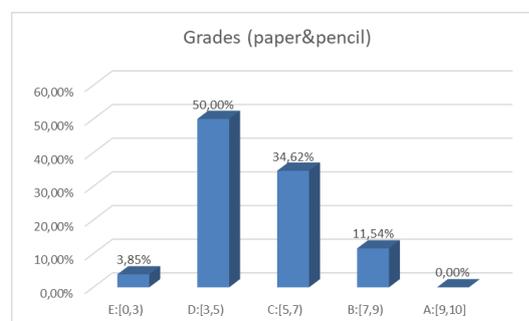

Figure 5: Distributions (paper and pencil) of (a) number of annotations produced by students (x-axis: number of annotations -interval, y-axis: percentage of students who produced annotations in the corresponding interval), and (b) grades obtained by the students (x-axis: grade intervals, y-axis: number of students who obtained a grade in each interval).

- Annotations with limited content (a simple underline, a succinct sentence or some lines of text, as shown in Figure 6a) and more elaborated annotations, but barely connected to the text and closer to the final critical analysis of the work were identified (Figure 6b).
- As shown in Figure 6, students usually adopted many different annotation styles, which indicates a lack of systematicity during annotation.
- As also reflected in Figure 6, superfluous annotations that were minimally related to the narratological principles and many critical aspects that were not considered were frequently observed.

(a)

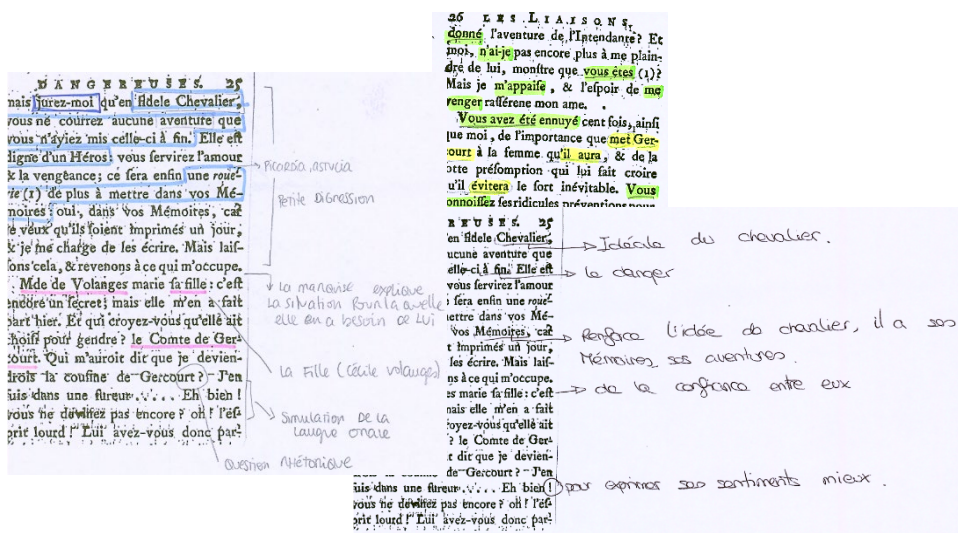

(b)

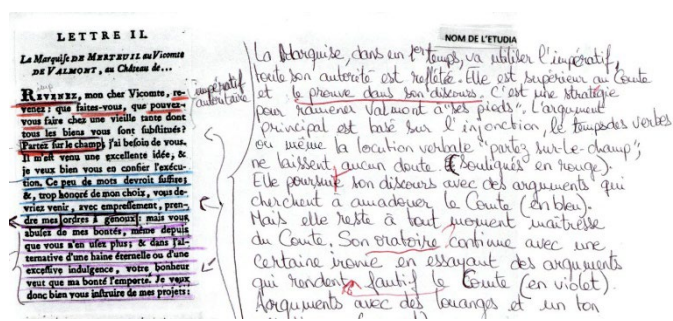

Figure 6: Examples of paper and pencil based annotations: (a) examples of succinct annotations; (b) example of elaborated annotations.

As a result, the grades for the activity were not satisfactory. Figure 5b shows the distribution of these grades, which were grouped into 5 categories (from E, the worst grade, to A, the best grade). Most of the students (approximately 54%) do not obtain the minimum grade to pass (C or higher). The average score (4.5, 95% CI [4.01, 4.99]) is below 5, which is the minimum grade to pass.

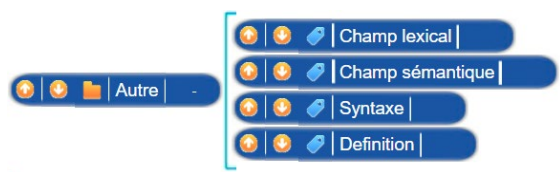

Figure 7: Concepts suggested by the students.

*4.2. Annotation with @note*

Concerning the activity with *@note*, 490 annotations were produced. A small number (42) of annotations were classified with concepts proposed by the students and grouped in the category *Autre* (*other*; Figure 7). Figure 8a summarizes the distribution of the 490 annotations produced by the 27 students who participated in the activity with @note. More than 51% of the students produced between 10 and 20 annotations. The average number of annotations produced by each student was 18.15 (95% CI [14.68, 21.62])

(a) (b)

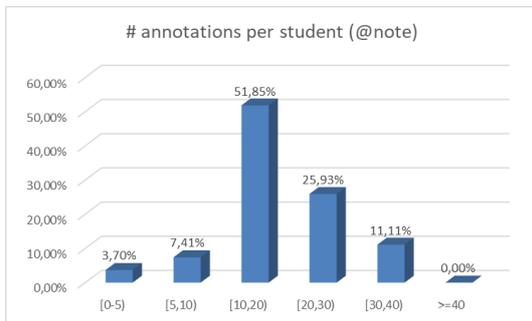 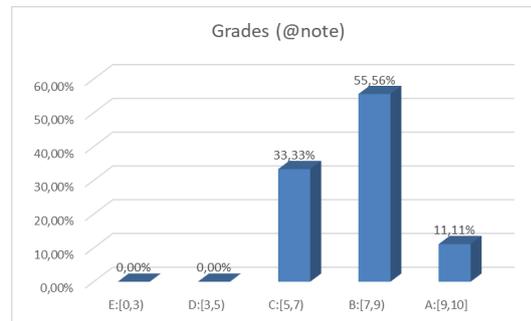

Figure 8: Distributions (@note) of (a) number of annotations produced by students, (b) grades obtained by the students (x-axis: grade intervals, y-axis: number of students who obtained a grade in each interval).

Figure 9: Examples of the elaborated @note based annotations

Concerning the *quality* of the annotations:
- The instructor observed a greater homogenization in the annotation process due to the annotation discipline introduced by the ontology-guided approach and imposed by the tool. The heterogeneity in the annotation modes disappeared in favor of a single annotation format based on a semantic tagging of the annotations and an elaboration in the form of an enriched-free text (Figure 9).
- The instructor observed a considerably more systematic critical annotation process: the students were forced to tag their annotations with concepts with a strong semantic charge and include strong arguments that support tagging, which contributed to a decrease in superfluous annotations.

Therefore, the activity outcomes were significantly better than those of the baseline, conventional paper-and-pencil activity as evidenced by the distribution of grades for the activity with *@note* shown in Figure 8b. The average score is 6.93 (95% CI [6.42, 7.42]). All students obtained sufficient qualification (minimum of 5) to pass the activity.

*4.3. Comparison*

The number of annotations produced by the students does not significantly differ between the paper and pencil and the *@note* annotation activities:

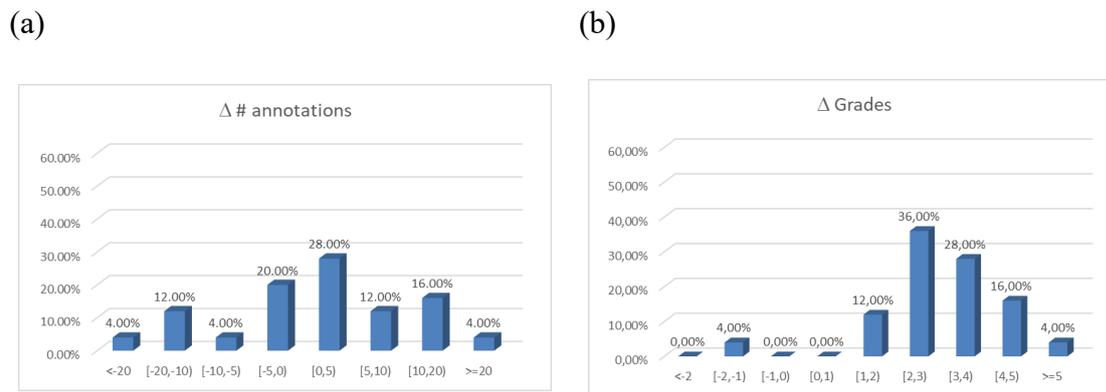

Figure 10: Distributions of (a) differences in annotations produced by each student (annotations in @note – annotations in paper and pencil activity), and (b) differences in grades (grades @note - grades paper and pencil).

- The average number of annotations are very similar in both activities (18 in the paper and pencil activity versus 18.15 in the *@note* one). Likewise, no statistically significant difference is observed between distributions in Figure 8 (Mann-Whitney U test: U = 339, p = .830).
- Figure 10a shows the distribution of the difference between the number of annotations produced with @note and the number of annotations produced with paper and pencil for the 25 students that participated in both activities. Although there is a tendency toward positive differences (40% negative versus 60% positive), this trend is not overwhelming. Consequently, the average of the difference in the number of annotations is 0.44 (95% CI [-3.7, 4.6]), and no statistically significant evidence is observed in favor of a non-zero median for the distribution (Wilcoxon signed rank test of the number of annotations for each student in both activities: Z = 150, p = .715).

   However, the final grades were significantly better with *@note*:
- The average score with *@note* is almost 2.5 points above the average score in the activity based on pencil and paper. The improvement is statistically significant (Mann–Whitney U test of the scores in both activities: U = 639, p = .000).
- This improvement is also evidenced when the analysis focuses on the improvement that is individually obtained by each student. Taking into account the 25 participants who engaged in both activities, Figure 10b shows the distribution of the differences in the grades between the *@note* activity and the paper and pencil activity. The majority of the students (84%) increased the grade obtained by more than 2 points. The average improvement does not significantly differ from the previously indicated value (2.44 points, 95% CI [1.87, 3.01]). The corresponding Wilcoxon signed rank test yields statistically significant evidence (Z = 316.5, p = .000) in favor of the improvement (median of the difference distribution is positive).

## 5. Discussion

The pilot's outcomes suggest that the ontology-driven approach can enhance the learning results with respect to an unguided, conventional paper-and-pencil based document annotation approach. A significant improvement in the quality of the annotations produced with the tool was observed with respect to those that were produced using the conventional method based on paper and pencil. The students produced annotations that were better situated and more reasonable than those observed in the paper and pencil activity, in which annotations that were irrelevant from a critical analysis perspective proliferated. Students applied the free text part of the annotation to justify the reason for the selected semantic concepts, argue how these concepts were reflected and developed in the text, and delve into the subtleties and complexities of the texts. Also, they omitted considerably fewer relevant aspects than the conventional paper and pencil approach (because the ontology already suggested to the students the aspects to seek and to be identified in the text). As a natural consequence of these improvements, the grades obtained with the tool were significantly better than those in the case based on paper and pencil.

Note the proposal by the students of the final concepts shown in Figure 7. The students had missed purely linguistic concepts such as "lexical field" and "semantic field", which are purely formal (as much as they can be "adjective" or "verb") and instead used more analytical and interpretative categories, such as "analogical writing" or "hyperbolic writing". In the context of the critical analysis of the text promoted by this activity, these needs enabled a diagnose of the lack of assimilation of the theory of analysis by some of the students who proposed and employed these concepts (an expected lack due to the knowledge-demanding nature of the critical literary annotation domain) rather than consideration of this lack as a weakness of the annotation ontology. Since *@note* enables students to propose concepts using this feature, it also partially illustrates the shortcomings of approaches such as the folksonomy-based approaches.

Concerning the internal validity of the results, we consider it unlikely that the observed enhancements were attributed to the potential carryover effects (fatigue or practice) of the *within-subject* design approach that was adopted in the pilot.

- Fatigue was avoided since the annotation activities (baseline, paper-and-pencil, and *@note* activities) were performed in a separated session, and in both cases, students had sufficient time to complete the work in a calm way (one week).
- Potential carryover effects due to practice were minimized. As the baseline activity was based on paper and pencil instead of another annotation tool, the previous practice of students with other annotation tools for critical annotation was explicitly avoided. In addition, the text in the *@note* activity was different from the text in the baseline activity and the two annotation activities obeyed two radically different annotation paradigms: unguided, free-style annotation in the baseline activity vs. ontology-guided annotation in the activity that involved *@note*. Therefore, we estimated the probability that the experience in the critical annotation gained by the students during the baseline activity had significantly influenced the realization of the activity with *@note* to be negligible.

## 6. Conclusions and Future Research

In this paper, we have described an ontology-driven approach to educational document annotation activities. This approach is particularly valuable for guiding students during the annotation of documents in complex learning domains, in which a considerable amount of expert knowledge is required. The approach has been implemented in *@note*, which is an annotation tool aimed at educational activities that require high cognitive skills. *@note* enables instructors to configure all aspects of an annotation activity. The central element of this process is the formulation of the annotation ontology. This ontology guides students during the annotation of the document since, instead of adding annotations in an unstructured way, they are forced to classify their annotations with the concepts of the ontology. The combination of free-text annotations with semantic annotations confers to *@note* a distinctive feature that is not present in any of the analyzed annotation tools. Therefore, each annotation must be clearly justified in relation to the annotation objectives reflected by the ontology, which promotes meta-reflective thinking in students. Consequently, as evidenced by the pilot, the learning results can be substantially improved.

We are currently exploring the application of the ontology-driven approach to contexts other than the critical annotation of literary texts. In particular, and in a context of teacher training for kindergarten education, we are applying *@note* for the interpretation of image-based children's storybooks [55]. We are also exploring the application of the approach to code reading activities in programming [56], as well as to activities of requirement analysis and transformation in software development, e.g., systematic development of compilers from specifications based on attribute grammars [57,58]. Another promising use of *@note* follows from its integration inside *Clavy*, a platform for building learning object repositories with reconfigurable structures [59–61]. In this setting, we utilize *@note* to enable healthcare students to annotate clinical images in radiologic reports that are integrated in radiology-specific collections of learning objects [62–64]. Finally, we consider it interesting to explore the possibility of simultaneous collaborative formulation of ontologies during the critical annotation process, involving more expert annotators, as well as the interplay between collaborative annotations by a group of more experienced students with the ontology-guided approach described in this paper.

**Supplementary Materials:** *@note* is available online at a-note.fdi.ucm.es.

**Author Contributions:** conceptualization, María Goicoechea-de-Jorge and Amelia Sanz and José-Luis Sierra; methodology, Antonio Sarasa and Mercedes Gómez and José-Luis Sierra; software, Joaquín Gayoso-Cabada and Mercedes Gómez and Antonio Sarasa and José Luis Sierra; validation, maría Goicoechea-de-Jorge and Amelia Sanz; formal analysis, Mercedes Gómez and Antonio Sarasa and José-Luis Sierra; investigation, Joaquín Gayoso-Cabada and María Goicoechea-de-Jorge and Amelia Sanz; resources, Joaquín Gayoso-Cabada and María Goicoechea-de-Jorge and Mercedes Gómez and Amelia Sanz and Antonio Sarasa and José-Luis Sierra; data curation, Joaquín Gayoso-Cabada and Mercedes Gómez and José-Luis Sierra; writing—original draft preparation, José-Luis Sierra; writing—review and editing, Joaquín Gayoso-Cabada and María Goicoechea-de-Jorge and Mercedes Gómez and Amelia Sanz and Antonio Sarasa and José-Luis Sierra; visualization, Joaquín Gayoso-Cabada and José-Luis Sierra; supervision, José-Luis Sierra; project administration, José-Luis Sierra; funding acquisition, María Goicoechea-de-Jorge and Antonio Sarasa and José-Luis Sierra.

**Funding:** This research was funded by the Spanish Ministry of Science, Innovation and Universities grant numbers TIN2014-52010-R (RedR+Human) and TIN2017-88092-R (CetrO+Spec) and by the Community of Madrid, grant number S2015/HUM-3426 (e-Lite CM).

**Acknowledgments:** The development of *@note* was initially supported by two awards of the Google's Digital Humanities Award Programs (2010 and 2011 editions).

**Conflicts of Interest:** The authors declare no conflict of interest. The funders had no role in the design of the study; in the collection, analyses, or interpretation of data; in the writing of the manuscript, or in the decision to publish the results.